# Her2 Challenge Contest: A Detailed Assessment of Automated Her2 Scoring Algorithms in Whole Slide Images of Breast Cancer Tissues

Talha Qaiser[a*], Abhik Mukherjee[b*], Chaitanya Reddy Pb[c], Sai Dileep Munugoti[c], Vamsi Tallam[c], Tomi Pitkäaho[d], Taina Lehtimäki[d], Thomas Naughton[d], Matt Berseth[e], Aníbal Pedraza[f], Ramakrishnan Mukundan[g], Matthew Smith[h], Abhir Bhalerao[a], Erik Rodner[i], Marcel Simon[i], Joachim Denzler[i], Chao-Hui Huang[j,k], Gloria Bueno[f], David Snead[l], Ian O Ellis[b], Mohammad Ilyas[b,m], Nasir Rajpoot[a,l]

[a] Department of Computer Science, University of Warwick, UK
[b] Department of Histopathology, Division of Cancer and Stem Cells, School of Medicine, University of Nottingham, UK
[c] Department of Electronics and Electrical Engineering, Indian Institute of Technology Guwahati, India
[d] Department of Computer Science, Maynooth University, Ireland
[e] NLP Logix LLC, USA
[f] VISILAB, E.T.S.I.I, University of Castilla-La Mancha, Ciudad Real, Spain
[g] Department of Computer Science and Software Engineering University of Canterbury, New Zealand
[h] Department of Statistics, University of Warwick, UK
[i] Computer Vision Group, Friedrich Schiller University of Jena, Germany
[j] MSD International GmbH, Singapore
[k] Singapore Agency for Science, Technology and Research, Singapore
[l] Department of Pathology, University Hospitals Coventry and Warwickshire, UK
[m] Nottingham Molecular Pathology Node, University of Nottingham, UK

Addressee for Correspondence:

**Prof Nasir Rajpoot and Talha Qaiser.**

Department of Computer Science, University of Warwick, UK;

Email: N.M.Rajpoot@warwick.ac.uk; T.Qaiser@warwick.ac.uk;

Running Title:

Automated Her2 Scoring Challenge Contest 2016

Conflicts of Interest: **None[1]**

Word Count: **4475**

* Both are first authors

## ABSTRACT

**Aims**

Evaluating expression of the Human epidermal growth factor receptor 2 (Her2) by visual examination of immunohistochemistry (IHC) on invasive breast cancer (BCa) is a key part of the diagnostic assessment of BCa due to its recognised importance as a predictive and prognostic marker in clinical practice. However, visual scoring of Her2 is subjective and consequently prone to inter-observer variability. Given the prognostic and therapeutic implications of Her2 scoring, a more objective method is required. In this paper, we report on a recent automated Her2 scoring contest, held in conjunction with the annual PathSoc meeting held in Nottingham in June 2016, aimed at systematically comparing and advancing the state-of-the-art Artificial Intelligence (AI) based automated methods for Her2 scoring.

**Methods and Results**

The contest dataset comprised of digitised whole slide images (WSI) of sections from 86 cases of invasive breast carcinoma stained with both Haematoxylin & Eosin (H&E) and IHC for Her2. The contesting algorithms automatically predicted scores of the IHC slides for an unseen subset of the dataset and the predicted scores were compared with the "ground truth" (a consensus score from at least two experts). We also report on a simple *Man vs Machine* contest for the scoring of Her2 and show that the automated methods could beat the pathology experts on this contest dataset.

**Conclusions**

This paper presents a benchmark for comparing the performance of automated algorithms for scoring of Her2. It also demonstrates the enormous potential of automated algorithms in assisting the pathologist with objective IHC scoring.

## Introduction

The adoption of image analysis in digital pathology has recently received significant attention due to the availability of digital slide scanners and the increasing importance of tissue-based biomarkers in stratified medicine [1]. Advances in software development and an upwards trend in computational capacity have also caused an upsurge of interest in digital pathology.

Breast Cancer (BCa) is the most commonly diagnosed cancer among women, and the second leading cause of death worldwide [2]. According to Cancer Research UK, the risk for women being diagnosed with breast cancer is 1 out of 8 in the UK, and approximately 11,600 women died from breast cancer in 2012 [3]. In routine diagnostic practice of BCa, tumour tissue is stained with Haematoxylin and Eosin (H&E) and then examined under the optical microscope for morphological assessment including grade. In addition, tissues are stained by immunohistochemistry (IHC) to evaluate biomarker expression for prognostic and predictive purposes. This conventional method of diagnosis by visual examination is considered accurate in most areas but is known to suffer from inter-observer and intra-observer variability in some areas such as diagnosis of atypical hyperplasia and reporting of histological grade [4–6]. Digital pathology offers significant potential for improvement to overcome the subjectivity and improve reproducibility.

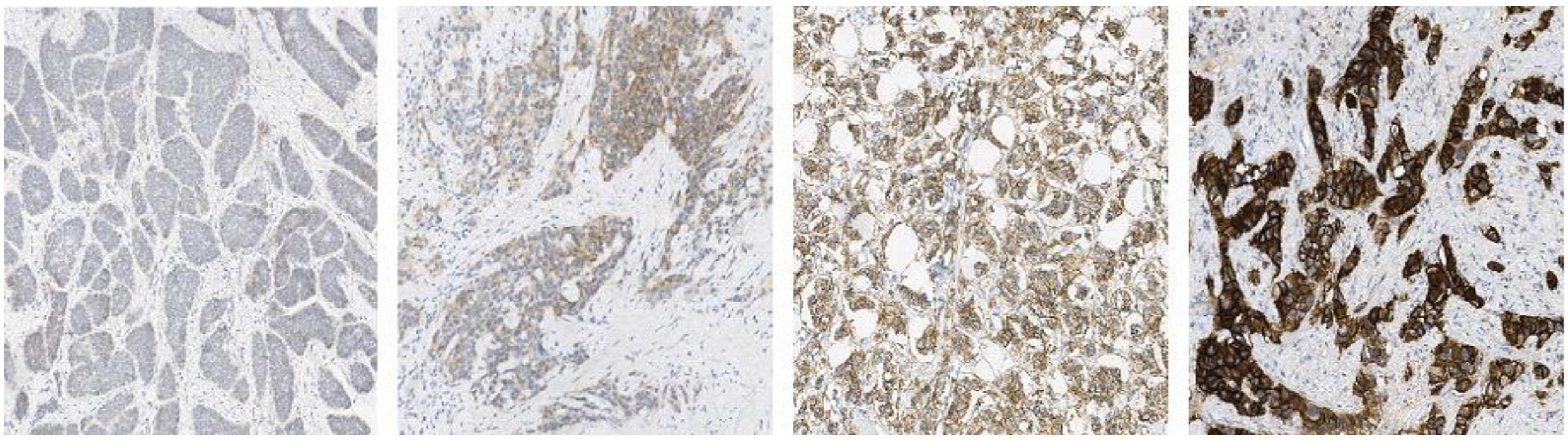

Fig 1: *Left to Right* - Examples of regions of interest (800$\mu m$ in height and the same in width) from WSIs scored 0, 1+ (negative), 2+ (equivocal), and 3+ (positive)

The human epidermal growth factor receptor 2 (Her2) gene is amplified in approximately 15-20% of breast cancers [7]. Gene amplification can also be identified through Fluorescence *In Situ* Hybridisation (FISH). Alternatively, since Her2 amplification results in increased protein expression, IHC may be used. Given the technical ease of performing IHC, it has become the preferred test and FISH is usually only performed when the IHC is equivocal. In practice, an expert histopathologist will report a score between 0 and 3+ and cases scoring 0 or 1+ are classified negative whilst cases with a score of 3+ are classed as positive. Cases with score 2+ are classified as equivocal and are further assessed by FISH to test for gene amplification. Examples of the four different Her2 scores (0 to 3+) are shown in Fig 1. A summary of recommended guidelines for Her2 IHC scoring criteria [7] is shown in Table 1.

| **Score** | **Cell Membrane Staining Pattern** | **Staining Assessment** |
|---|---|---|
| 0 | No membrane staining or incomplete membrane staining in $<$ 10% of invasive tumour cells (0+) OR faint/barely perceptible or weak incomplete membrane stainaing in $>$ 10% of tumour cells (1+) | Negative |
| 1+ | | Negative |
| 2+ | A weak to moderate complete membrane staining is observed in $>$ 10% of tumour cells OR strong complete membrane staining in $\leq$10% of tumour cells | Borderline (Equivocal) |
| 3+ | A strong (intense and uniform) complete membrane staining is observed in $>$ 10% of invasive tumour cells | Positive |

Table 1: Recommended Her2 scoring criteria for IHC stained breast cancer tissue slides [7]

Historically, up to 20% of the Her2 IHC results may contain inaccuracies [8] due to variations in the technical quality and the subjective nature of scoring. Although adoption of Her2 guidelines and recommendations [7], have served to improve standards in Her2 testing, there remain challenging cases especially with Her2 scores deemed borderlines between categories.

Automated IHC scoring of Her2 carries promise to overcome the existing problems in conventional methods. Automated scoring methods are not prone to subjective bias and can provide precise quantitative analysis which can assist the expert pathologist to reach a reproducible score.

The Her2 Scoring Contest, documented in this paper, was organized by the University of Warwick, the University of Nottingham and the Academic-Industrial Collaboration for Digital Pathology (AIDPATH) consortium (www.aidpath.eu). It was held in conjunction with the Pathological Society of Great Britain and Ireland meeting in Nottingham (June 2016) to provide a platform for researchers to assess the performance of computer algorithms for automated Her2 scoring on IHC stained slides. This paper provides an overview of the automated methods for Her2 scoring as presented at the contest and a *Man vs Machine* comparison of the degree of agreement among histopathologists and the automated methods for Her2 scoring. This may be considered as an initial step towards the development of a reliable computer-assisted diagnosis tool for Her2 scoring of digitised BCa histology slides.

## Materials and Methods

## Ethics

The ethics approval was by Nottingham Research Ethics Committee 2 [Approval No: REC 2020313]; R&D reference (N) 03HI01.

## Image Data Acquisition and Ground Truth

The histology slides for this contest were scanned on a Hamamatsu NanoZoomer C9600 enabling the image to be viewed from a ×4 to a ×40 magnification, making the process comparable to a clinician's standard microscope. Generally, WSIs are gigapixel images stored in a multi-resolution pyramid structure where the highest resolution is ×40. The contest dataset entailed 172 whole slide images (WSI) extracted from 86 cases of invasive breast carcinomas and included both the H&E and Her2 stained slides. The actual Her2 scoring is normally done on the IHC stained slides whilst the H&E slides assist the expert pathologist to identify the areas of invasive tumour and discriminate these from areas of *in situ* disease. Fig 2 shows an example of the two types of WSIs (with a corresponding zoomed-in region of interest) from the contest dataset.

The ground truth (GT) was taken from the clinical reports issued on the cases at a tertiary referral centre for breast pathology (Nottingham University Hospitals, NHS Trust). At this centre, each case had been reported or reviewed by at least 2 specialist consultant histopathologists as part of their routine practice (preliminary reporting and MDT review). The centre provides regular internal quality control for Her2 assessment for immunohistochemistry runs and regularly contributes and participates in the UK NEQAS (National External Quality Assessment Scheme) for immunocytochemistry and *in situ* hybridisation (ICC & ISH).

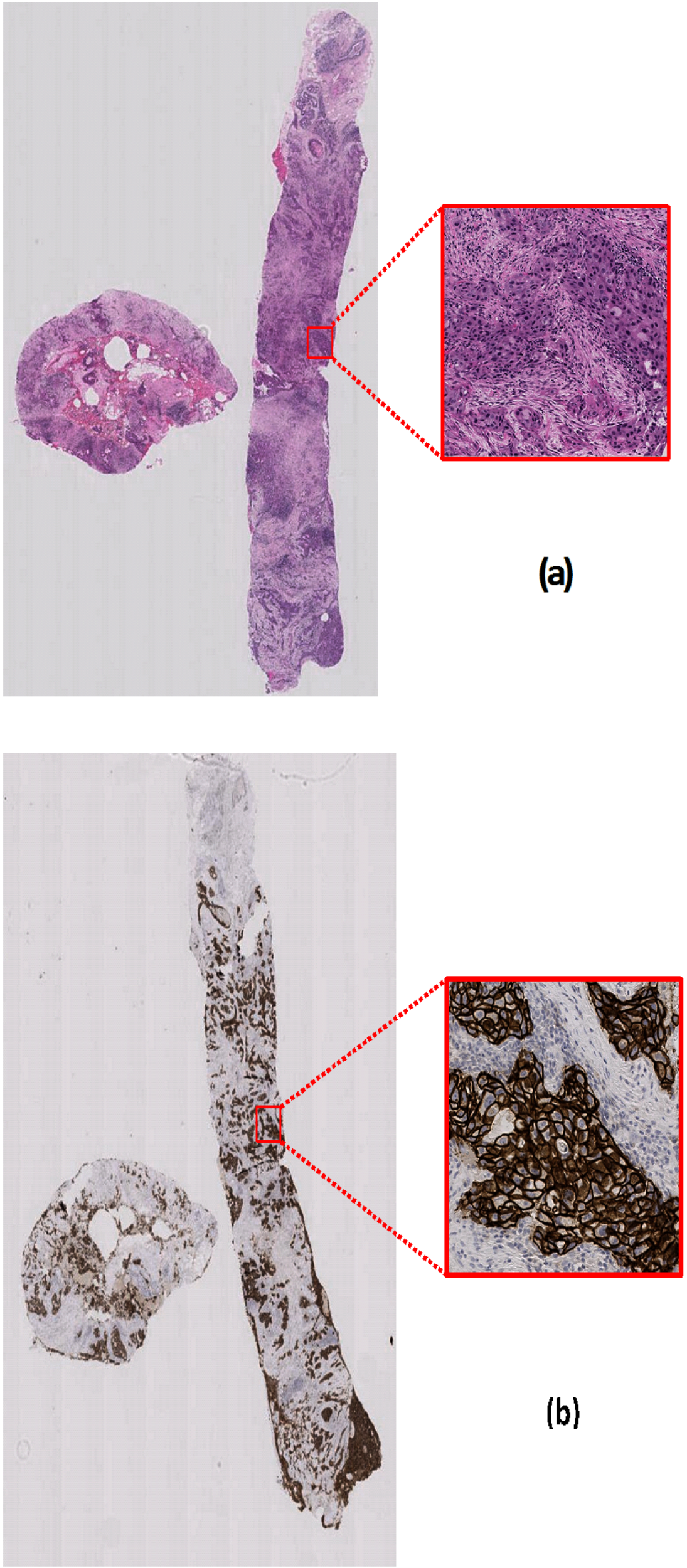

Fig 2: An example WSI along with a zoomed-in cross-sectional area showing the tumour region (a) H&E stained slide (b) IHC stained slide

## Contestants

A total of 105 teams from more than 28 countries registered to access the training dataset before the end of the registration deadline. By the end of submission deadline (off-site contest), a total of 18 submissions from 14 teams were received for evaluation. The organizers provided an opportunity to each of the 14 teams for presenting their approach in the contest workshop and 6 teams chose to present. For the *Man vs Machine* contest, we received the markings from 4 pathologists. The contest website was reopened for new submissions after concluding the workshop. Further details regarding various stages of the contest is described in Supplementary Material A.

## Evaluation

The performance of each submitted algorithm was evaluated based on three criteria: 1) agreement points, 2) weighted confidence, and 3) combined points. Each assessment criterion has a separate leader-board.

The evaluation criteria were rationalised according to the clinical significance and implications of Her2 IHC scoring as follows: in everyday clinical practice, for a score of 0 and 1+: No Herceptin is offered to the patient; for 3+ score, Herceptin is offered. For an IHC 2+ score, a FISH test is performed; if positive (i.e.) there is evidence of gene amplification and Herceptin is offered while for a negative result, it is not offered. The evaluation considers the impact of erroneous classification. For example, a score of 0/1+ being interpreted as 3+ or vice versa is a serious error while a 2+ scored as 0/1+ denies a few patients of valid treatment; a score of 3+ for a 2+ case bypasses the FISH test and may erroneously treat few cases (which would have been FISH negative) with toxic drugs while a score of actual 3+ downgraded to a 2+ calls for additional expense of FISH testing but the end result will probably be the same and hence this should not be regarded as that serious an error. These have been summarised in Table 2.

For agreement points, a penalty method was employed whereby each erroneous prediction is penalised with respect to its deviation from the GT as shown in Table 2 (a). It can be envisaged that the agreement points may end in a tie, where the accumulative points of two or more teams may be the same. To resolve the tie, a bonus criterion was devised as shown in Table 2 (b), where the decision was made on the percentage of cells with complete cell membrane staining (PCMS) regardless of the intensity. The bonus points were primarily introduced for score 2+ and 3+ as they attain more clinical significance. For the IHC score 1+, 1 bonus point was awarded if there was an accurate prediction of the IHC score and PCMS $< 3\%$, while 3 bonus points were awarded if there was an accurate prediction of the IHC score and PCMS $> 3\%$ but the predicted PCMS value only deviated $\pm 2\%$ from the GT. For the IHC scores 2+ and 3+, 5 bonus points were awarded if there was an accurate prediction of the IHC score and PCMS only deviated $\pm 5\%$ from the GT. Similarly, 2.5 bonus points were awarded for score 2+ and 3+, if there was an accurate predication of IHC score and PCMS only deviated $\pm 10\%$ from the GT.

| | | Predicted Score | | | |
|---|---|---|---|---|---|
| **Ground Truth** | **Score** | **0** | **1+** | **2+** | **3+** |
| | **0** | 15 | 15 | 10 | 0 |
| | **1+** | 15 | 15 | 10 | 0 |
| | **2+** | 2.5 | 2.5 | 15 | 5 |
| | **3+** | 0 | 0 | 10 | 15 |

(a)

| Ground Truth Score | Percentage of cells with complete cell membrane staining (PCMS) | |
|---|---|---|
| **0** | 0 | 0 |
| **1+** | 1 (PCMS < 3%) | 3 (PCMS ± 2) |
| **2+** | 5 (PCMS ± 5) | 2.5 (PCMS ± 10) |
| **3+** | 5 (PCMS ± 5) | 2.5 (PCMS ± 10) |

(b)

Table 2: (a) Agreement points for predicted calls of ground truth (GT), (b) Bonus point criteria, when PCMS lies in certain range of the GT value of the PCMS.

The weighted confidence was devised to measure the credence of the predicted score by the submitted algorithm. The criteria to measure the weighted confidence $w_c$ were distinct for both truly and wrongly classified cases. In cases where the predicted Her2 score $p_s$ matched with the GT with higher confidence $c$, the weighted confidence amplified the confidence value for true prediction whereas wrong predictions with high confidence were penalized accordingly, as given in equation (1). This kind of assessment is important for the development of an interactive diagnostic module. The confidence value may indicate those cases or regions where further examination by the experts may be required before concluding the final Her2 score.

$$w_c = \begin{cases} \frac{2c-c^2}{2} & if \ \ p_s = GT \\ \frac{-c^2+1}{2} & otherwise \end{cases} \qquad (1)$$

The third assessment criterion is a combination of both agreement points and weighted confidence based evaluations. The combined points were calculated by taking the product of two assessment criteria for each case.

## Results

## Contest Leaderboards

Comprehensive results comprising all the submissions for automated methods are shown in Table 3. The teams in were ranked with respect to the combined-point based assessment with bonus points. For the off-site contest, the total possible points were 420 (28 cases with a maximum of 15 points each) whereas for weighted confidence, the maximum points were 28, 1 for each case. The top three ranked teams with respect to point based assessments were Team Indus, MUCS-1, MUCS-2 whereas according to weighted confidence assessment the top ranked teams were VISILAB, FSUJena, MTB NLP. The combined results rank the top three team in the following order: VISILAB, FSUJena and Huangch. The performance of top-ranked teams including bonus points and the trend for total points (without the bonus points) can be seen in Fig 3. MUCS-1, MUCS-3, CS_UCCGIP and MTB NLP achieved equal points but MUCS-1 secured more bonus points as their PCMS was more accurate as compare to remaining counterparts. Similarly, Team VISILAB and Rumrocks ended up in a tie where both teams attained equal points but the VISILAB method was more precise in predicting PCMS. Comprehensive tables for all three leaderboards are available for download from the contest website.

## Summary of Proposed Automated Methods

Most of the automated methods (described in Supplementary Material B) applied a supervised patch based classification approach to solve this problem. The most common pipeline was based on three main components: 1) pre-processing including the methods to identify the regions of interest for patch generation, 2) classification based on handcrafted or neural network learned features, and 3) post-processing techniques to aggregate the Her2 score at WSI level and to estimate the PCMS. Deep learning, especially Convolutional Neural Network (CNN) based approaches dominated as 8 out of top 10 methods were based on CNN. The majority of the CNN architectures (Team Indus, MUCS-(1-3), MTB NLP, VISILAB, RumRocks, FSUJena) were inspired from the state-of-the-art deep neural networks [9,10].

| Rank | Team | Affiliation | Points | Points + Bonus | Weighted Confidence | Combined |
|---|---|---|---|---|---|---|
| 1 | **VISILAB** | Universidad de Castilla-La Mancha | 382.5 | 404.5 | 23.552 | 348.041 |
| 2 | **FSUJena** | Computer Vision Group, University of Jena | 370 | 392 | 23 | 345 |
| 3 | **HUANGCH** | Bioinformatics Institute, Singapore | 377.5 | 391.5 | 22.622 | 335.77 |
| 4 | **MTB NLP** | NLP Logix, LLC | 390 | 405.5 | 22.937 | 335.737 |
| 5 | **VISILAB (Density)** | Universidad de Castilla-La Mancha | 377.5 | 391 | 21.878 | 322.067 |
| 6 | **Team Indus** | Indian Institute of Technology Guwahati | 402.5 | 425 | 18.451 | 321.414 |
| 7 | **UC-CSSE-CGIP Group** | University of Canterbury, New Zealand | 390 | 395 | 21.07 | 316.05 |
| 8 | **MUCS – 3** | Computer Science, Maynooth University | 390 | 411 | 20.434 | 300.813 |
| 9 | **HERcules** | University of Oxford | 360 | 380 | 20.572 | 295.633 |
| 10 | **MUCS – 2** | Computer Science, Maynooth University | 385 | 413 | 19.51 | 290.171 |
| 11 | **Rumrocks** | Department of Statistics, University of Warwick | 382.5 | 395 | 19.649 | 277.705 |
| 12 | **TissueGnostics** | TissueGnostics GmbH, Austria | 365 | 366 | 17.78 | 266.41 |
| 13 | **Team Indus (Stainsep)** | Indian Institute of Technology Guwahati | 332.5 | 345.5 | 18.451 | 250.715 |
| 14 | **MUCS – 1** | Computer Science, Maynooth University | 390 | 416 | 16.765 | 248.876 |
| 15 | **HersRockers** | Indian Institute of Technology Guwahati | 320 | 330 | 17.318 | 223.007 |
| 16 | **VIP-UGR** | University of Granada | 305 | 322.5 | 15.41 | 211.748 |
| 17 | **TartanSight** | Computational Biology, CMU | 230 | 230 | 15.148 | 159.745 |
| 18 | **Cancer_Detector** | Indian Institute of Technology Kanpur | 255 | 260 | 12.994 | 138.962 |

Table 3: A summary of results of all three assessment criteria for the automated Her2 scoring contest, ordered by the combined points criterion.

In pre-processing and patch extraction stage, most of the teams followed the conventional thresholding techniques with a combination of morphological operators. These techniques are computationally less expensive and generally work well as background regions lack any texture contents in contrast with other tissue components. The MUCS-(1-3), MTB NLP, VISILAB and FSUJena manually probe the regions of interest through some calibration or customized methodologies. These methods aimed to pick the best possible regions for training their algorithm, generally without affecting the testing phase. To segment tissue regions, the RumRocks team implemented a deconvolutional neural network (DCNN) and a 2D CNN being for selection of patches based on their texture. The Huangch team performed mean filtering and stain normalization by using the control tissue intensity values to calibrate the stain colour intensity as a pre-processing step.

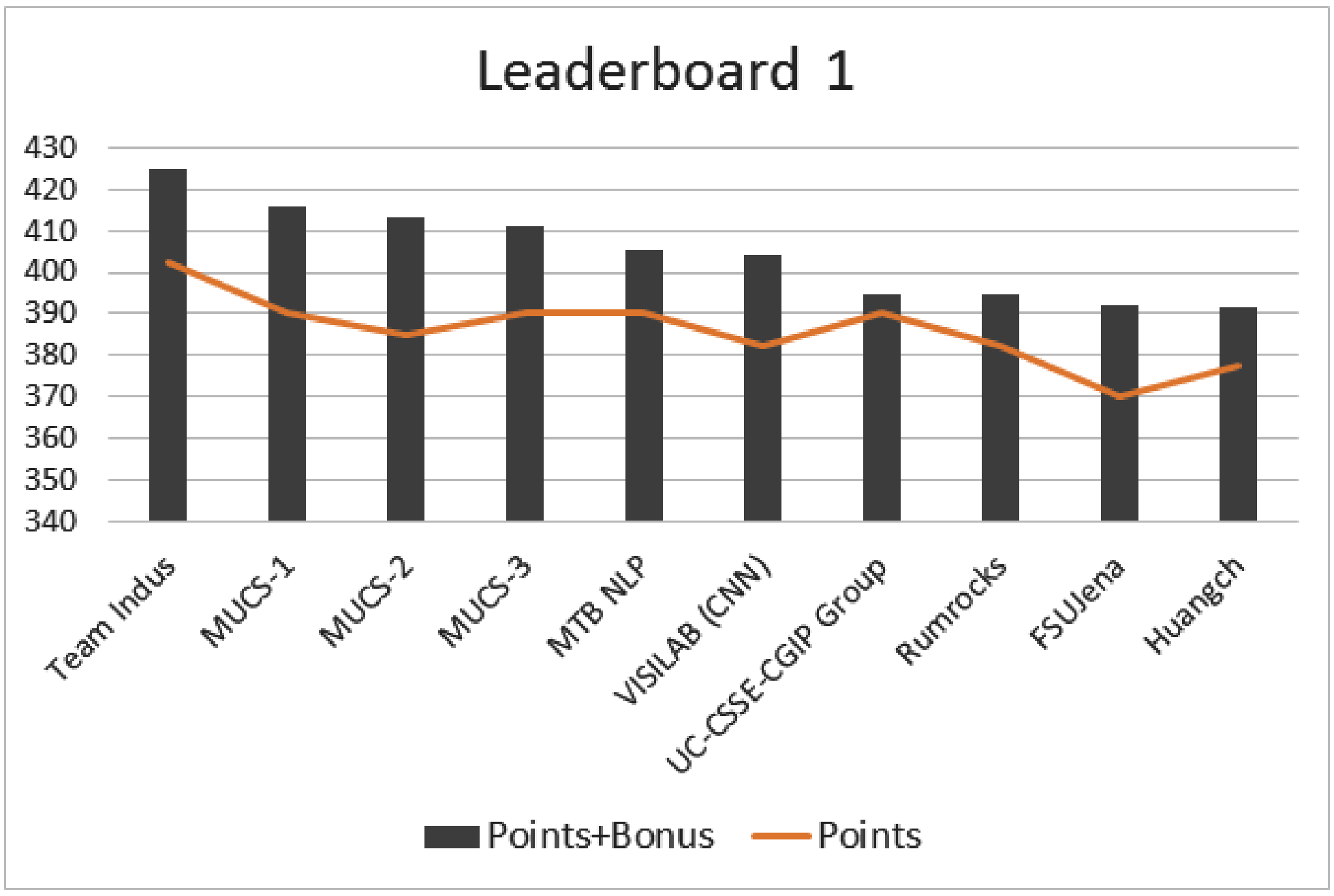

Fig 3: Combined results for top ranked teams with respect to agreement and bonus points.The trend shows the significance of correclty predicting the percantage of cell membrane.

In the second step, most of the teams (specifically top-10) employed deep learning approaches whereas other teams like CS_UCCGIP and Huangch derived handcrafted characteristic curves and employed standard machine learning approaches. Team Indus used a combination of data-driven and handcrafted features. They incorporated the average control tissue intensity value along with learned features maps before passing them to the fully connected layers. Some of the top-ranked teams deployed variants of Alexnet [9] and GoogLeNet [10] for predicting the Her2 score. The FSUJena team computed the bilinear features after retrieving activations from convolutional layers of the AlexNet. The derived activations contain the learned feature maps representing a $d$-dimensional $w \times h$ spatial grid. This approach enables them to perform their analysis on top of the learned features maps from CNN. In combination with standard approaches for data regularization, MTB NLP and RumRocks trained multiple models. The final Her2 score and PCMS was estimated by averaging over all the models. Additionally, a wide range of data augmentation and regularization techniques were employed to overcome the overfitting issues. As in practice, the standard data augmentation techniques such as affine transformations (e.g. rotation, flip, translation), random cropping, blurring and elastic deformations were applied to train the network. MUCS-2, MTB NLP and RumRocks broadly used the data augmentation techniques to assist the network to generalize well on unseen data.

In the final stage of pre-processing and predicting the PCMS, most of the teams employed standard image processing and machine learning approaches on top of the results attained from the last step. A Random Forest classifier was trained by MTB NLP to produce the final class probabilities and to estimate the PCMS. FSUJena simply used the mean tumour cell percentage seen in the training set for a particular class as an estimate. Team Indus used both IHC and H&E stained slides to estimate the PCMS by using standard image processing approaches like contour detection, thresholding and morphological features. All the remaining teams limited their analysis to only IHC stained images. All the submissions used high-magnification images (10× or above) except MUCS and Rumrocks who used images from low resolution for selection of ROIs.

## ***Man vs Machine* Event**

### **Organization**

One way of evaluating the automated algorithms for IHC (Her2) scoring is to perform comparative analysis of the assessment of expert pathologists and automated methods for a handful of cases as compared to the scores for those cases as agreed by at least two consultant breast-pathologists (GT). On the day of contest workshop, we organized an event called as *Man* *vs Machine*. The main aim of this event was to analyse the performance of automatic methods and to explore the disagreements among conventional and automatic methods. This kind of analysis can lead us to a more sophisticated protocol for automatic Her2 scoring and to overcome the inter- and intra-observer agreements that can be found in normal practice.

The analysis between the expert's agreement and the evaluation of the automatic Her2 scoring method was performed with a subset (15 cases) of the off-site test dataset. For this event, we set up an online webpage for the pathologists. The webpage enabled the experts to load and navigate (including pan and zoom) through the WSI of those cases. Both IHC (Her2) and H&E stained digital images were made available to mimic the conventional scoring environment. We requested the expert pathologists on the contest day at PathSoc 2016 to score each case by providing the Her2 score, PCMS and a confidence value.

## ***Man vs Machine* Results Comparison**

Table 4 summarizes the overall evaluation scores achieved by each participant for this event. Each table entry gives the cumulative score for all 15 cases, which indicates the overall performance. The agreement-points based assessment was used to evaluate the performance for this event. In total, we received 4 responses from expert pathologists and as shown in Table 4, we ranked the top 6 submissions including the top 3 automated methods. From submitted responses, three participant pathologists reported themselves as 'Consultant Pathologist' and one as 'Trainee Pathologist' and all three of them marked breast pathology as a subspecialty.

As can be seen in Table 4, one of the automated methods slightly outperformed the top-performing participant pathologist. These results point to the potential significance of automated scoring methods and the recent advancements in digital pathology. It's worth mentioning that automated Her2 scoring algorithms submitted in this contest are not ready to deploy in their current form, as they will require extensive validation on a significantly large-scale data and also plenty of input from experts to prepare the GT on the larger data.

| Rank | Team Name | Score | Bonus | Score + Bonus |
|---|---|---|---|---|
| **1** | Team Indus | 220 | 12.5 | 232.5 |
| **2** | Expert 2 | 210 | 20.5 | 230.5 |
| **3** | VISILAB | 212.5 | 15 | 227.5 |
| **4** | MUCS-1 | 205 | 20.5 | 225.5 |
| **5** | Expert 1 | 185 | 10 | 195 |
| **6** | Expert 3 | 180 | 13 | 193 |

Table 4: Summary Results for the *Man vs Machine* event. The evaluation was carried out according to the contest criteria as described in Evaluation Section.

Table 5 shows pooled data for Her2 scoring among the three top-ranked automated methods and the scores from three participant pathologists and comparison with the GT. The Table 5 was determined for the 15 cases selected from the off-site contest dataset. On the basis of Her2 scores, a 100% agreement with the GT was observed for score 3+ among the participant pathologists and the automated methods. For the scores of 1+ and 2+, there were disparities between the GT and the new scores. In all cases bar one, for both man and machine, the error resulted from overcalling the score. Thus, for the score 1+, on 6/9 (67%) were overcalled as 2+ by humans whilst 4/9 (44%) were overcalled by the machine algorithms. For the score of 2+, 7/15 (46%) were overcalled as 3+ by humans whilst machines overcalled 1/15 (6%) as 3+ and 1/15 (6%) was undercalled as 1+. Clinically, score of 2+ is critical, as in routine practice cases of score 2+ are recommended to go through FISH testing. It’s equally important to avoid predicting the score 2+ as 1+ or 0, cases such erroneous prediction will deny the further assessment of Her2. As it can be seen in Table 5, none of the cases with score 2+ was misclassified by the participant pathologists as either 1+ or 0 whereas for one of the case an automated method wrongly predicted a score of 2+ as 1+.

| Case | Ground Truth | FISH Results | Expert 1 | Expert 2 | Expert 3 | Team Indus | Visilab | MUCS-1 |
|---|---|---|---|---|---|---|---|---|
| 1 | 2+ | Negative | 3+ | 2+ | 2+ | 2+ | 2+ | 2+ |
| 2 | 0 | - | 0 | 1+ | 1+ | 1+ | 1+ | 0 |
| 3 | 3+ | - | 3+ | 3+ | 3+ | 3+ | 3+ | 3+ |
| 4 | 0 | - | 1+ | 1+ | 1+ | 0 | 1+ | 1+ |
| 5 | 1+ | - | 2+ | 1+ | 2+ | 1+ | 2+ | 1+ |
| 6 | 3+ | - | 3+ | 3+ | 3+ | 3+ | 3+ | 3+ |
| 7 | 2+ | Borderline amplified | 3+ | 3+ | 3+ | 2+ | 2+ | 2+ |
| 8 | 2+ | Negative | 3+ | 2+ | 3+ | 2+ | 3+ | 2+ |
| 9 | 3+ | - | 3+ | 3+ | 3+ | 3+ | 3+ | 3+ |
| 10 | 3+ | - | 3+ | 3+ | 3+ | 3+ | 3+ | 3+ |
| 11 | 1+ | - | 1+ | 1+ | 2+ | 0 | 1+ | 1+ |
| 12 | 2+ | Positive | 2+ | 2+ | 3+ | 2+ | 2+ | 2+ |
| 13 | 1+ | - | 2+ | 2+ | 2+ | 2+ | 2+ | 1+ |
| 14 | 2+ | Negative | 2+ | 2+ | 2+ | 2+ | 2+ | 1+ |
| 15 | 0 | - | 0 | 1+ | 0 | 0 | 1+ | 0 |
| 16 | 2+ | Borderline amplified | - | - | - | 0 | 1+ | 2+ |
| 17 | 2+ | Negative | - | - | - | 2+ | 2+ | 2+ |
| 18 | 2+ | Positive | - | - | - | 2+ | 1+ | 2+ |
| 19 | 2+ | Borderline amplified | - | - | - | 2+ | 2+ | 2+ |
| 20 | 1+ | - | - | - | - | 1+ | 1+ | 1+ |
| 21 | 1+ | - | - | - | - | 1+ | 1+ | 2+ |
| 22 | 0 | - | - | - | - | 1+ | 0 | 1+ |
| 23 | 1+ | - | - | - | - | 0 | 1+ | 1+ |
| 24 | 1+ | - | - | - | - | 0 | 1+ | 2+ |
| 25 | 3+ | - | - | - | - | 3+ | 3+ | 3+ |
| 26 | 0 | - | - | - | - | 1+ | 0 | 1+ |
| 27 | 0 | - | - | - | - | 0 | 0 | 1+ |
| 28 | 0 | - | - | - | - | 0 | 0 | 0 |

Table 5: Combined matrix for agreement among the three experts and the top three automated methods based on agreement points against the GT scores for 15 cases in the *Man vs Machine* event. The boderline case 7 was deemed negative and cases 16,19 were deemed positive for treatment decision (based on the HER2:CEP17amplification ratio for Her2 over-expression: 1.96, 2.1 and 2.07 respectively).

Most of the incorrect predictions by the participant pathologists were found to be in cases where there was considerable heterogeneity. Two such examples are shown in Fig 4 (a-d). In tumour cells of Her2 score 2+, a pattern of weak to moderate complete membrane staining is observed whereas for score 3+, an intense (uniform) complete membrane staining is observed. Estimating the complete membrane staining is a difficult and highly subjective process especially for score 2+ and 3+, as it is extremely hard to pick up subtle differences in the morphological appearance for those cases.

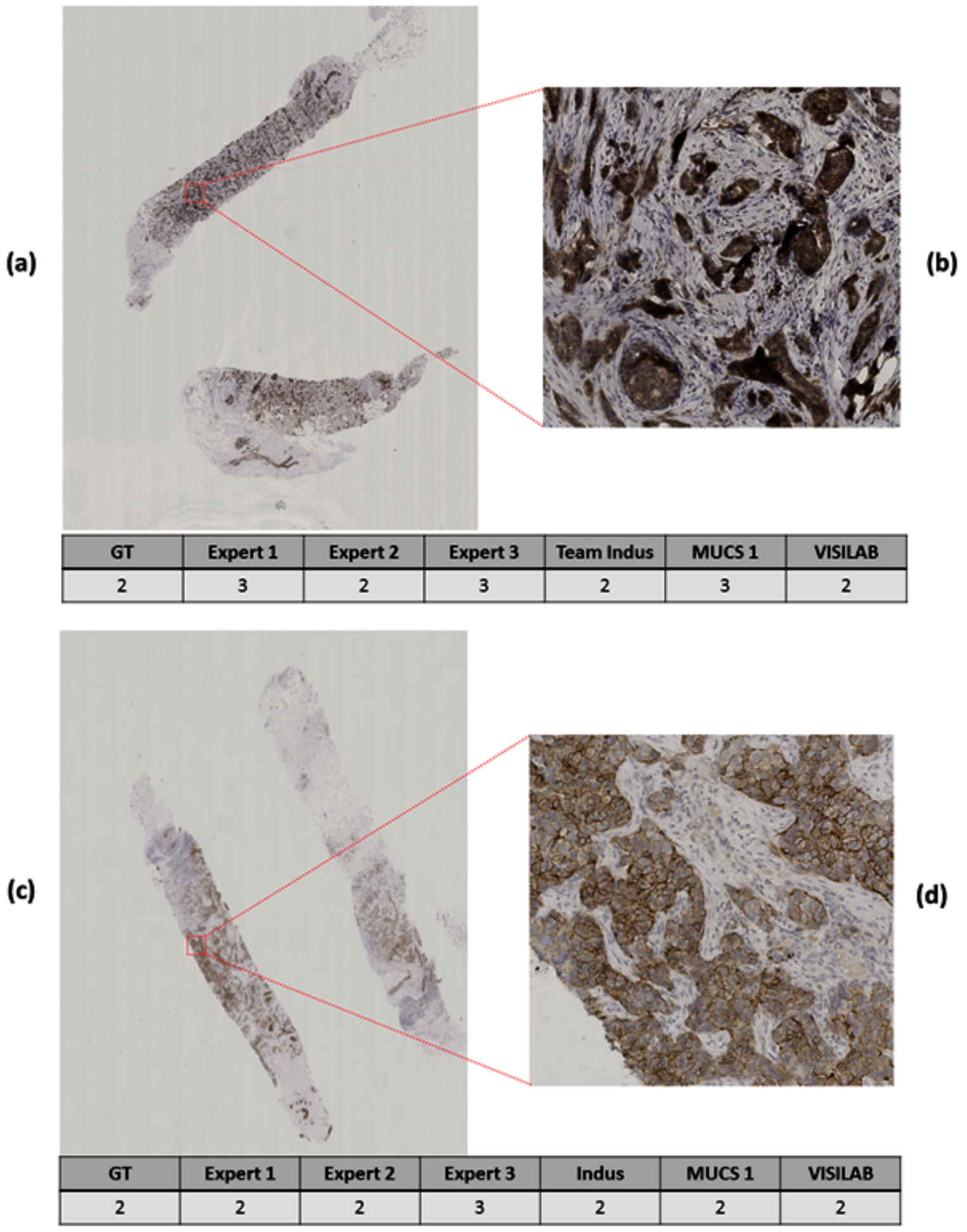

| GT | Expert 1 | Expert 2 | Expert 3 | Team Indus | MUCS 1 | VISILAB |
|---|---|---|---|---|---|---|
| 2 | 3 | 2 | 3 | 2 | 3 | 2 |

| GT | Expert 1 | Expert 2 | Expert 3 | Indus | MUCS 1 | VISILAB |
|---|---|---|---|---|---|---|
| 2 | 2 | 2 | 3 | 2 | 2 | 2 |

Fig 4: Examples showing IHC stained WSIs (a & c) and zoomed-in cross sectional area (b & d) with corresponding Her2 GT scores marked by expert pathologists and predictions from top 3 automated methods.

## Discussion

A major aim of organizing this contest was to provide a platform for computer scientists and researchers to contribute and to evaluate the performance of their computer algorithms for automated IHC scoring of Her2 in images from BCa tissue slides. Automated scoring can significantly overcome the subjectivity found due to varying standards adopted by different diagnostics labs. There is a current wealth of literature [11,12] using individual platforms (both freely and commercially available) for digital analysis of Her2 in BCa. This, however, was the first comparison of platforms and algorithms and provides a pilot for independent comparison of computing algorithms for Her2 assessment on a benchmark dataset. The contest highlights the wealth of potential carried by Artificial Intelligence (AI) techniques for assessment of IHC slides.

The contest "training dataset" was deliberately selected in a way that it contained a reasonable number of cases from all Her2 scores bearing in mind the need for the training algorithms to learn features for each score. For the test dataset (both off-site and on-site), the GT was withheld at the time of image evaluation. Results showed that the automated analysis performed comparable to histopathologists. Many of the algorithms achieved high accuracy – often close to the maximum. Our main objective was to analyse the performance of algorithms based on clinical relevance and hence the three particular evaluation criteria described above were chosen. It may be possible that other assessment criteria may influence the ranking of comparative results.

The data from the Man vs Machine comparison showed that, reassuringly, all participants (whether human or computer) correctly identified cases with GT score of 3+. This means that no-one in the category would have been denied treatment. Similarly, for the cases with a score of 0 or 1+, although there was some over-calling, this never exceeded 2+ and thus none would have received treatment without further testing. The most problematic category was, not unexpectedly, cases with a score of 2+; in both human and machine evaluations. If overcalled as 3+, the FISH negative subset would be over-treated. The GT information for the FISH results were not released to the participants as the contest was aimed just at comparing interpretation of Her2 IHC results. Hence, most of the automated algorithms aimed at predicting the equivocal cases as 2+. Table 5 incorporates the FISH results for all the cases that were marked as 2+ in the test data GT (including *Man vs Machine* dataset). From *Man vs Machine* cases (15 in total), a score of 2+ (subsequently FISH negative) was overcalled by the machine as 3+ in just one instance (VISILAB). In contrast, on three occasions (subsequently FISH negative) the participant pathologists overcalled the score 2+ as 3+. Moreover, for the remaining test dataset (13 cases), on three instances the score of 2+ (subsequently FISH positive) were erroneously predicted as either 1+ and 0 by the automated algorithms. Overall, the results indicate that further fine-tuning will be required for 2+ cases with AI. While it is encouraging that automated Her2 scoring algorithms may have sufficient potential, as direct comparison to human diagnosis, it is probably worthwhile to reflect that the number of pathologists actually joining the contest was small (only four) and it would have been better to compare the pathologist's assessment of the slides on a reporting microscope rather than a computer for a fairer comparison to real life practice.

Conventionally, expert pathologists often switch back and forth between the IHC and H&E slides to map the invasive tumour regions for estimating the percentage of complete membrane staining. Beside one of the participants (Team Indus), most of the algorithms reported in this paper have avoided the use of H&E slides, though one cannot rule out the use of H&E slide for automatic detection of DCIS regions. In addition, the task of predicting the PCMS is extremely subjective, as the expert has to make estimation on the basis of physical appearance of the stained invasive tumour region. The semi-automated methods could provide a comprehensive quantitative analysis on selected region of interest to assist the experts in estimating the PCMS and Her2 score, especially on borderline cases. As Her2 immunoscoring relies not only on intensity but the completeness of membrane positivity, automated scoring may be helpful as demonstrated by Brügmann *et al*. [13] who proposed scoring of Her2 based on an algorithm evaluating the cell membrane connectivity.

This study shows that automated IHC scoring algorithms can provide a quantitative assessment of morphological features that can assist in objective computer-assisted diagnosis and predictive modelling of the outcome and survival [14]. We have demonstrated the potential significance of digital imaging and automated tools in histopathology. In the context of breast histopathology, whereby almost all the invasive tumour cases are considered for Her2 testing, an automated or semi-automated scoring method has potential for deployment in routine practice. Despite of all these advancements, several challenges remain for the AI algorithms to be optimised and to become the part of routine diagnosis. It is worth noting that serious optimization will be needed for automated methods while processing a whole-slide image. Some methods required more than three hours per case, which, in the "real world" of diagnostic service delivery is not feasible. Another limitation of this contest was that the image data were collected from a single site using a single scanner. A potential extension would be to collect data from multiple pathology laboratories with Her2 scores marked by different experts and images scanned using a variety of different machines. This would also test the differences inherent in staining quality that may affect such procedures. Such enhancements could significantly overcome the overfitting to one particular dataset that may occur in the automated scoring methods. In moving across systems, other laboratories for example, have acknowledged the challenges in reaching the optimum Aperio algorithm parameters to provide results that were equivalent to those of the 'Automated Cellular Imaging System' (ACIS) or 'Cell Analysis System' (CAS 200) quantitation systems [15], which are fully automated environments for detecting cells based on intensity characteristics and handcrafted features found in IHC stained images. Therefore, there is a need to learn across comparative systems for which the current study provided a valid starting point. Also, the study highlights the need of dialogue between histopathologists and informaticians to understand correct identification of tissue compartments relevant for assessment, correct morphology (normal vs *in situ vs* invasive) and stromal stain *vs* tumour stain. Algorithms will also need to be trained to the natural acceptable variation in staining hues and intensities (intra and inter-laboratories) to work effectively during routine practice.

All cases with score 2+ are routinely recommended for further FISH testing to validate Her2 overexpression at the gene level. It would be an added advantage if the automated methods could be trained with FISH GT to predict the final outcome and the potential for automated algorithms in calling the actual final Her2 status with reproducible accuracy could be demonstrated. For this, a larger series with 2+ cases alone with FISH data would need to be tested. Indeed, there have been promising other studies that indicate that automated image analysis for Her2 instead of manual assessment may reduce the need for supplementary FISH testing by up to 68% [16]. In a diagnostic setting, this would significantly reduce costs and turn-around time. During the last decade, IHC staining has become ubiquitous in pathology labs around the world and the role of IHC evaluation in a high-throughput setting becomes key for IHC based companion diagnostics. Other possible extensions of digital pathology could be to automate the overexpression of the programmed death 1 (PD-1) receptor and its ligand (PD-L1), to evaluate anaplastic lymphoma kinase (ALK) protein and proto-oncogene tyrosine-protein kinase ROS1 in lung cancers [17]. The AI based algorithms would be more effective if IHC staining and scoring methods were treated as a composite assay [18][19]. The varying staining protocols and scoring parameters may restrain the effectiveness of AI based automated scoring algorithms including the Her2 scoring but with sufficiently variable data from different centres, AI algorithms could be trained to overcome that problem.

This contest provides a baseline for computer science and computational pathology researchers for automated/semi-automated scoring and computer-assisted diagnosis (CAD) tools to assist the pathologists in daily routine analysis. The contest is now over but the registration and the web-portal will remain open for future participants to make novel contribution in automated Her2 scoring.

**Acknowledgements**

The first author (Qaiser) acknowledges the financial support provided by the University Hospital Coventry Warwickshire (UHCW) and the Department of Computer Science at Warwick. The VISILAB team (Pedraza and Dr. Bueno) and UNOTT (Prof. Ilyas and Dr. Mukherjee) acknowledge financial support from the European Project AIDPATH (no.: 612471). http://aidpath.eu/. The MUCS team wishes to acknowledge John McDonald and Ronan Reilly for their valuable contributions to the research, and acknowledge financial support from Science Foundation Ireland (SFI) under grant no. 13/CDA/2224 and an Irish Research Council (IRC) Post Graduate Scholarship. Co-first author, Dr. Mukherjee would also like to thank the NIHR and Pathological Society of Great Britain and Ireland for support. We are also grateful to Dr. Nicholas Trahearn for his input in deriving the weighted confidence evaluation measure.

## **Supplementary Material**

A: Contest Format

B: Description of Automated Methods

# Supplementary Material A

## Contest Format

The contest involved four stages, as described below.

**Stage 1:** Release of the Training Data In first stage, a training dataset comprising 52 cases were released to the registrants on April 24, 2016 through a secure website portal[1]. The dataset consisted of IHC and H&E stained images and the ground truth (GT). The GT score and percentage cells with complete membrane staining for the released training dataset can be seen in Table 1. At this stage, most of the details regarding contest (like tasks, contest rules, contest forum details etc) were already posted to the contest website and the registered teams started their work on algorithms for Her2 scoring. The registration process remained open for five weeks. We also created a social-forum (Google group) for the participants to share their queries and to communicate with the organizers.

**Stage 2:** Release of Off-Site Test Data A dataset comprising 28 cases were selected for off-site testing. This test dataset was released on May 17, 2016 and consisted of IHC and H&E stained WSIs without the GT information to ensure a fair evaluation. Source code for performance assessment in both MATLAB and Python languages were also released to the registrants. The registrants were given more than a month after releasing the test data to finalize and submit their scoring methods for announced tasks.

**Stage 3:** Submission of Results (Off-Site) The deadline for submission of results for the test dataset was set to be June 21, 2016, a week before to the contest workshop. Each team had to submit results in a comma-separated values (CSV) file along with a maximum 2-page summary of their algorithms, a description of experimental setup, and some preliminary results. The participants were advised that the CSV file should contain the predicted Her2 score, the confidence value for predicted score and the percentage of cells with complete cell membrane staining (PCMS) for each WSI in the test dataset. Each registrant was allowed to submit up to three sets of results. The submitted results were evaluated but results were not announced until the contest workshop was held.

**Stage 4:** Contest Workshop. The contest workshop was conducted in Nottingham in conjunction with the annual meeting of the Pathology Society of Great Britain and Ireland on June 29, 2016. The contest workshop covered three main events: a) a brief talk from the organizers and the participants where 6 teams were invited for a small presentation to give an overview about their approaches and experiments, b) announcement of the comparative results of algorithms for both off-site, and c) announcement of results for the *Man vs Machine* comparison as a part of the on-site contests. The remaining 6 cases (of the 86) were used for an on-site competition (although they were released one day before the contest workshop due to the computational requirements of some of the automated algorithms and their results are not discussed here). The complete tables of results are available on the contest website[2].

| **Case** | **Ground truth** | **FISH Results** | **Percentage cells with complete membrane staining irrespective of intensity** |
|---|---|---|---|
| 1 | 0 | N/A | 0% |
| 4 | 2 | Negative | 60% |
| 6 | 2 | Positive | 40% |
| 9 | 3 | N/A | 70% |
| 11 | 3 | N/A | 90% |
| 12 | 1 | N/A | 5% |
| 13 | 0 | N/A | 0% |
| 14 | 1 | N/A | 1% |

[2] http://www.warwick.ac.uk/TIAlab/Her2Contest/

| | | | |
|---|---|---|---|
| 15 | 1 | N/A | 2% |
| 16 | 1 | N/A | 5% |
| 18 | 0 | N/A | 0% |
| 19 | 3 | N/A | 30% |
| 22 | 3 | N/A | 90% |
| 24 | 1 | N/A | 5% |
| 25 | 2 | Negative | 30% |
| 26 | 2 | Positive | 50% |
| 27 | 3 | N/A | 80% |
| 29 | 0 | N/A | 0% |
| 30 | 3 | N/A | 90% |
| 32 | 1 | N/A | 3% |
| 33 | 3 | N/A | 100% |
| 34 | 1 | N/A | 2% |
| 35 | 3 | N/A | 90% |
| 36 | 2 | Positive | 100% |
| 38 | 3 | N/A | 90% |
| 39 | 0 | N/A | 0% |
| 40 | 2 | Positive | 60% |
| 46 | 0 | N/A | 0% |
| 47 | 1 | N/A | 5% |
| 48 | 2 | Positive | 20% |
| 49 | 2 | Positive | 30% |
| 50 | 2 | Positive | 50% |
| 52 | 0 | N/A | 0% |
| 55 | 2 | Positive | 70% |
| 57 | 0 | N/A | 0% |
| 58 | 1 | N/A | 5% |
| 61 | 3 | N/A | 90% |
| 63 | 2 | Borderline amplified | 70% |
| 65 | 1 | N/A | 2% |
| 66 | 0 | N/A | 0% |
| 67 | 2 | Positive | 30% |
| 68 | 0 | N/A | 0% |
| 70 | 0 | N/A | 0% |
| 73 | 0 | N/A | 0% |
| 74 | 2 | Positive | 10% |
| 79 | 1 | N/A | 5% |
| 82 | 3 | N/A | 80% |
| 83 | 3 | N/A | 100% |
| 84 | 3 | N/A | 70% |
| 86 | 1 | N/A | 3% |
| 87 | 0 | N/A | 0% |
| 88 | 1 | N/A | 5% |

Table 1: The ground truth score for 52 cases from the training dataset with percentage of cells with complete membrane staining. The boderline case 63 was deemed negative and the amplification ratio for Her2 over-expression was 1.92.

# Supplementary Material B

## Related Work on Automated IHC Scoring

Automated image analysis is observed as a solution [1,2] to overcome the inter- and intra-observer variations found in conventional assessment of tissue slides. Hence, the automated scoring of routine H&E and IHC stained slides has received huge interest in recent years. In literature, several classical machine learning approaches [3–5] have been presented but recently deep learning based approaches have been profoundly employed for H&E and IHC histology image analysis [6,7].

In literature, a wide range of handcrafted features was proposed for IHC scoring algorithms [4,5]. For instance, Choudhury *et al*. [8] proposed an averaged threshold measure (ATM) for scoring of digitized images of IHC stained tissue microarrays. A set of arbitrary chosen thresholds was selected, whereby an optimal threshold using the ATM is used for calculating the percentage of stained area. The proposed ATM statistic presented as a generalization of the HSCORE [9] statistic for scoring IHC slides. Reyes-Aldasoro *et al*. [10] presented an alternative approach for automated segmentation of microvessels in IHC tumor slides. For segmentation, distinguishing hues of stained vascular endothelial nuclei and tissue regions were explored to extract the seeds for a 'region-growing' model. Their post-processing of segmented microvessels from CD31 immunostaining contained three steps, closing morphological objects from tumour margins, combining isolated objects, and splitting objects into individual vessels with having multiple lumina. Although the thresholding approaches perform well on a specific dataset, they are likely to fare not as well on an unseen dataset as distinctive hues can be significantly varying. A potential reason of such variation lies in staining process, as the histology slides normally stained at different occasions with inconsistent concentrations often exhibit large variations in colour and appearance. Such differences in slide preparation make the colour and morphological appearance of tissue components more unpredictable.

Kuse *et al*. [11] used local isotropic phase symmetry measure as a significant feature for beta cell detection and lymphocytes. By calculating the peak of median phase energy after stain normalization but due to heterogeneous appearance and often-clumped structure makes nuclei segmentation a non-trivial task. Khan *et al*. [5] used stain quantization for the scoring of Estrogen Receptor (ER) and Progesterone Receptor (PR) by determining the amount of chromatin material and protein content from IHC stained WSIs. Ali *et al*. [12] used astronomical algorithms for the scoring of ER on IHC stained images of breast cancer. However, in this contest the classical machine learning approaches have been outperformed by deep learning approaches. Most of the published algorithms are based on different approaches with different dataset whereas this contest provides a platform where participants can develop and validate the performance of their algorithms on same dataset.

## Description of Automated Methods

The concise description of automated methods employed by top-ranked teams are described below.

## Team Indus

In this approach, a deep convolutional neural network (CNN) was employed for predicting the Her2 score whereas for estimating the percentage of complete membrane staining, a set of handcrafted morphological features were extracted from H&E and IHC stained slides.

**Pre-processing**: The patches with average edge strength lies higher then certain threshold were selected for training CNN.

**Her2 Score Prediction**: The presented CNN architecture contains five convolutional layers, one concatenation layer with following two fully connected and one classification layer. After each convolution and fully connected layer, a ReLu activation was performed whereas for classification layer a softmax activation was placed. After convolution layers a concatenation layer was positioned. The concatenation layer combines the activation maps from the convolution layers and the average control tissue intensity for the corresponding WSI from which the patches were originated. The weights for training CNN were initialized using H&E normal initializations [13] and updated using mini batch gradient descent (learning rate = 0.00015, weight decay = $10^{-6}$, Nesterov momentum = 0.95, batch size = 32). The CNN was trained over 41K patches generated each of size 224x224 from 52 training WSIs for 65 epochs.

During testing, the trained network assigned a score to each patch of a WSI and to aggregate the patch scores into a single Her2 score following criteria was proposed. Let $n_0$, $n_1$, $n_2$ and $n_3$ be the number of patches scored as 0, 1+, 2+ and 3+ respectively and N be the total number of patches generated from a WSI.

*If $n_3/N > 0.08$:*

*predict 3+*

*else if $n_2/N > 0.4$:*

*predict 2+*

*else if $n_1/N > 0.14$:*

*predict 1+*

*else:*

*predict 0*

**Percentage of Complete Membrane Staining (PCMS)**: To estimate the PCMS, first tumor regions were identified by extracting the morphological features from tumor and normal regions of H&E images.

After performing stain normalisation [14], the hematoxylin channel was extracted to segment the nuclei using Otsu thresholding. Further, nuclei contours were fit around each individual structure and filtered on basis of area and eccentricity. This resulted in tumor identification regions by detecting the tumour nuclei based on their roundness and size. In order to estimate the extent of membrane staining, the morphological features were extracted from an IHC image. In addition, a contagious chicken-wire pattern was observed for complete membrane stained regions whereas other tissue components result in a fragmented/broken-up skeleton. Further, by filling holes in the chicken-wire skeleton and by measuring similarity with the original binary image the extent of membrane staining was estimated.

The PCMS is estimated by calculating the ratio between extent of membrane staining and tumor identification regions as given below.

$$PCMS = \frac{\text{extent of membrane staining}}{\text{tumor identification regions}}\ x\ 100$$

**MUCS**

In this submission, the well-known neural networks Alexnet [15]and GoogLeNet [16] were adapted by adjusting the layer specific parameters, such as kernel size, stride, and padding. There were three submissions from the MUCS team with two submissions using Alexnet (MUCS-1 and MUCS-2) and one using GoogLeNet (MUCS-3).

**Training**: The training dataset was obtained by hand-picking the regions of interest from 52 training IHC images that were considered to contain the most representative samples from each class. The regions were selected from the low resolution (0.625×) and mapped to the highest resolution (40×) whereupon each region was divided into 128 x 128 pixel patches.

The MUCS-1 trained network had four output classes with corresponding Her2 scores from 0 to 3+. MUCS-2 and MUCS-3 had an additional output class for the background. The background class contained the regions with texture having only a weak appearance of nuclei (without blueish or brownish colour). The training dataset for MUCS-2 was extended by data augmentation (rotation and mirroring) and by adding the hand-picked regions from test images (without knowing the classification of the slide it originated from). The total patches for MUCS-1, MUCS-2 and MUCS-3 were 29000, 319000 and 33500, respectively. The training images were divided between actual training data (75%) and validation data (25%). For all three submissions, the base learning rate was set to 0.001, and the learning rate was dropped every one-third of the maximum iterations by a factor of 10 ($\gamma$=0.1). The mean pixel value was subtracted from the training dataset.

**Classification**: For testing, the common regions from H&E and IHC were selected at a low resolution and those regions were mapped to maximum resolution to generate the patches for testing. Further, adaptive thresholding was applied to each patch, with an offset of 10, to produce a binary image. If the proportion of ones in the binary image was smaller than a factor of 0.9, then patch was classified with the trained neural network model, otherwise the patch was marked as background and therefore did not require classification. The Her2 score for a WSI was determined using the classified patches as follows:

- Score 3+, if patches with class 3 was greater than or equal to 10% of total patches
- Score 2+, if patches with class 2 was greater than or equal to 10%, or patches with class 3 was between 1% and 10%, of total patches
- Score 1+, if patches with class 1 was greater than or equal to 10% of total patches
- Score 0, otherwise

The confidence value for each WSI was calculated by averaging the confidence values of each patch. PCMS was calculated by summing the number of Score 3+ and 2+ patches and dividing the sum by total number of patches (excluding the background) as

$$PCMS = 100(n_2 + n_3)(\sum_{s=0}^{3} n_s)^{-1} \qquad (2)$$

where n is the number of patches given score s, s ∈ {0,1,2,3}

**MTB NLP**

A CNN was trained to predict the Her2 score for 128 x 128 patches of the WSI. Furthermore, as a post-processing step, a Random Forest model was trained to aggregate an estimated Her2 score and percentages of cell membrane for the WSI.

**Pre-Processing**: In the first, tissue regions were manually annotated at 40× by drawing regions from IHC stained slide images. A class label was assigned to each annotated region that corresponds to WSI GT score. In total, there were 272 annotated regions with an average size of 800 x 800.

**Patch Classification**: The architectures similar to Alexnet [15] and VGG-16 [17] were trained to predict the Her2 scores but the results were only submitted for the architecture similar to Alexnet. The annotated regions were separated at case level by using 207 regions for training and the remaining 65 for validation. Each patch was randomly flipped and rotated to increase the training dataset and a dropout layer [18] was positioned to prevent the overfitting. The model was trained with a total of 8,575,000 patches and with cross entropy loss.

For testing, each WSI split in to non-overlapping patches of 128 x 128 and fed in to the trained network for predictions. Further, connected component analysis based approach was carried out to merge 128 x 128 patches into clusters. For each of the class labels, aggregate metrics were computed for the WSI that captured the percent of the slide pixels.

**Aggregate Her2 Score and Percentage**: To predict the Her2 score and PCMS process the aggregated metrics were computed during the patch classification. These metrics were used as predictors for a Random Forest classifier that produces that final class probabilities for each of the WSI. The same process was repeated using a Random Forest regressor to estimates for the percentage of cells that contained staining.

The 5-fold cross validation was done on all of the 52 training images. In each fold, all of the test images were scored and the predicted scores and percentage estimates were averaged over all folds to produce the final estimates.

**VISILAB**

In this method, the state-of-the-art GoogLeNet [16] was implanted to predict the Her2 score and the percentage of complete cell membrane.

**Data Preparation**: A handcrafted dataset was built. For this purpose, a set of representative patches of the four Her2 scoring classes were extracted from the ground truth WSIs. Additionally, an extra class was employed to collect background samples. These extracted patches from training WSIs were 68 x 68 pixels size each. A total of 5750 patches were selected with an average of 1150 patches per class. The dataset was further split in to training (75%) and validation (25%) dataset.

**Training**: Among several state-of-art CNNs, GoogLeNet was finally selected for submission according to the results on validation dataset. The prepared dataset was used for training, by selecting 0.01 as base learning rate, with a decreasing policy over 50 epochs, using the Stochastic Gradient Descent.

**Classification**: The algorithm takes a WSI and applies a grid technique to obtain the corresponding patches, with a similar size than the ones from the training dataset. These are later classified with the trained model, whose output is a class prediction and a percentage of confidence over that decision.

**Her2 Scoring**: Once every single patch is classified, a single class score is provided for the WSI. The decision rule takes into account the percentage of patches that belongs to each class (omitting the background, which was treated as a separate class) using the following criteria: starting from class 3+ to class 1+, the first one to achieve at least 10% of patches is chosen as final decision. Regarding the percentage of cells with full membrane staining, an expert rule was developed. The knowledge basis came from the alternative techniques that were also developed, such as the calculation of the staining density for the nuclei. As a result, a relationship between the classes percentage distribution and the percentage of membrane cell staining was discovered.

**UCCSSE**

This method is based on characteristics curves, a novel feature descriptor for predicting the Her2 score. In pre-processing phase, five regions of interest (ROI) were extracted from each WSI, each of size 1800 x 1200 at 20×. The only condition for selecting the ROIs was to select those regions that should not contain more than 30 % pixels as background.

The segmentation step consists of identifying the tissue portion including the IHC stained membrane. The selected ROIs were first segmented in HSB and CIELab colour spaces. In addition, some colour filters and neighbourhood masks were used to segment the connective tissues and fat lobules that should be separated before calculating the PCMS.

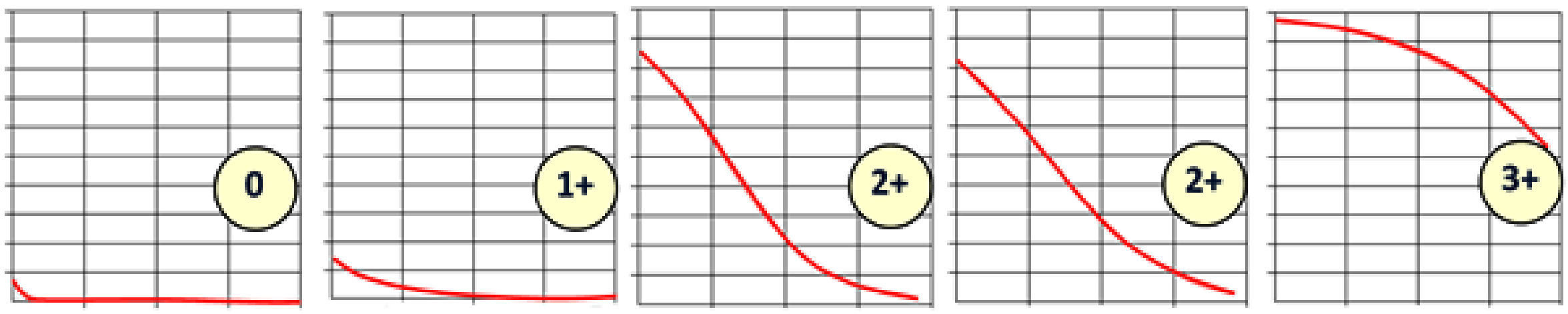

Fig 1: Characteristics curves and the corresponding Her2 score. The x-axis denotes range of the saturation value whereas y-axis denotes the calculated percentage from saturation limits. The predicted Her2 scores are also shown for each curve.

The essential part for the classification algorithm was the extraction of a characteristics curve for selected ROIs. The percentage-saturation characteristics curve was generated by varying the saturation limits from [0.1, 1] to [0.5, 1] in 20 steps by keeping the hue fixed. To plot the characteristics curve the percentage of stained region was calculated for each step by taking the ratio between segmented pixels to the number of pixels in an ROI. The characteristics curves have high discriminative appearance as shown in Fig 1. The curve always represents a smooth polynomial curve that can be accurately modelled using a cubic polynomial (best fit).

It was also observed during experimental analysis that when the Her2 score is 1+, the starting region of the curve always starts above the 10% mark depicting the presence of weak and incomplete membrane staining of regions. For 3+ score, the curves were lying above the 30% mark that shows the existence of an intense and uniform membrane staining areas.

**RumRocks**

In this approach, the two-dimensional (2D) CNN [15,19] models were trained for pre-processing and classification. First, as pre-processing step each WSI processed using deconvolution neural network (DCNN) and following by a $CNN_1$ to select the desired patches. Furthermore, the selected patches were processed through a $CNN_2$ to predict the Her2 score and the PCMS.

**Patch Selection**: A low resolution representation of a WSI (0.3125×) was selected and passed through a DCNN to segment the tissue components. Next, the detected regions were divided into patches with only condition that selected patches should contain 50% or more region from area of interest. The subsampled patch coordinates were translated to 10× resolution for further processing. The $CNN_1$ trained to accept or reject a subsampled patch based on its morphological appearance. The overview of neural network architectures are as given below

$$DCNN_1 = \{D_1, D_1, \dots \dots, D_6 - U_1, U_2, \dots \dots, U_5 - C_{2D}3\backslash 1 - Sg\}$$

$$CNN_1 = \{D_1, D_1, \dots \dots, D_7 - Reshape - FC - Sg\} \qquad (3)$$

The notation of the architecture is as follows, $D_1$ represents the down-sampling convolutional whereas $U_1$ represents the up-sampling convolutional layer with 3 as kernel size and 1 as stride. After convolutional operations batch normalization, ReLu and max pooling operations were applied. $FC$ represents fully connected layers and $Sg$ represents sigmoid function.

**Classification:** For predicting the Her2 score and PCMS, a $CNN_2$ with combination of residual layers [20] was employed. The batch dimensions were exploited in order to feed in multiple patches from the same WSI simultaneously. Instead of combining the prediction of individual patches through averaging or aggregating metrics, a tensor was reshaped to a vector once the spatial size has been significantly reduced and forward it through a 1D convolution layer. The overview of architecture $CNN_2$ is given below

$$CNN_2 = \{C_{2D}3\backslash 1 - resB_1, \dots\dots, resB_7 - flatten - C_{1D}1\backslash 1, - FC_1 - FC_2 - Sg\ \} \qquad (4)$$

The CNN models were trained using the mean squared error loss function and the Adam stochastic gradient decent optimization method with initial learning rate of $10^{-3}$. The learning rate was reduced every 15,000 iterations by a factor of 1.5 and trained each network for between 200,000 – 300,000 iterations. The average was calculated for each networks prediction to form an ensemble based score.

**FSUJena**

The algorithm for automated Her2 scoring was based on Alexnet [15] CNN. In this method, an activation matrix was extracted after convolution layers to compute the bilinear filters for predicting the Her2 score and PCMS.

At the first, ROIs were manually probed and patches of size 227 x 227 were randomly extracted at 20×. The pre-trained version of Alexnet was used from ImageNet dataset for further training on contest dataset. For each patch in the training dataset, an activation matrix was extracted after convolutional layers. The activations can be represented as a tensor $x \in \mathbb{R}^{w \times h \times d}$ comprised of $d$-dimensional vectors in a $w \times h$ spatial grid. The bilinear features [21,22] were further computed as the Gramian $G$ matrix by summing up dyadic products along the spatial dimensions: $G = \sum_{i,j} x_{i,j}\ , x_{i,j}^T$. The matrix $G$ contains the second-order statistics of the CNN features and have been found to be extremely useful for fine-grained recognition tasks. Then the square root and $L_2$normalization of $G$ were employed to increase the numerical stability of further processing steps [22]. To differentiate among four scoring classes a multi-class logistic regression was used. It was also observed that using a pre-trained network on ImageNet dataset is also beneficial to avoid the overfitting issues. In preliminary results the bilinear features approach outperformed the conventional CNN activations.

For testing a WSI and to predict the Her2 score, an average was calculated for all the random crops patches. To predict the PCMS the mean tumour cell percentage seen in the training set of for a particular class as an estimate.

**Huang's Method (Huangch)**

In this approach, a range of handcrafted features extracted from the IHC stained slides after performing the stain deconvolution. The handcrafted features were then fed in to a model of multi-class AdaBoosted decision trees.

**Sampling**: At the first, control tissue was extracted to developed a pseudo color space for stain deconvolution [23] to obtain the two staining vectors. Further, mean filtering was performed to record the local maximal points. The patches were selected from each WSI on the basis of local maximal points as they were representing the strongest Her2 stained over-expression signals

**Feature Extraction and Classification**: A combined but numerically independent features vector space constructed by including Gabor Filtering, Features of Fractal Dimension by Differential Box-Counting [23], multi-wavelet methods, histogram statics methods, grey-level (over all colour channels) co-occurrence based methods [24,25] etc.

For predicting the Her2 score and the PCMS, a model of multi-class AdaBoosted decision-trees was employed to map the features vector of each patch to a predicted value. This model is known as Stagewise Additive Modelling using a Multi-class Exponential [26] loss function (SAMME). The model composed by a series of decision-trees by assigning a weight to each decision-tree. Whereas while training, a pool of decision-trees generated and after each iteration the best decision-tree was selected with its corresponding weight. After certain iterations, a group of decision-trees was selected for testing phase.